\documentclass[runningheads]{llncs}
\usepackage{graphicx}

\usepackage{hyperref}
\usepackage{times}
\usepackage{soul}
\usepackage{url}
\usepackage[utf8]{inputenc}
\usepackage{graphicx}
\usepackage{amsmath}
\usepackage{booktabs}
\usepackage{algpseudocode}
\usepackage{fixltx2e}
\usepackage{lipsum}
\usepackage{amssymb,enumitem}
\usepackage{cases}

\usepackage{multicol}
\usepackage{multirow}
\usepackage{color}
\usepackage{xcolor}
\usepackage{booktabs}
\usepackage{graphicx}
\usepackage{nicefrac}
\usepackage{amsmath}
\usepackage{array}
\newcolumntype{P}[1]{>{\centering\arraybackslash}p{#1}}

\usepackage{wasysym}
\usepackage{graphicx}
\usepackage{fancyhdr}
\usepackage{url}
\usepackage{color}
\usepackage{marvosym}

\usepackage{float}

\usepackage{caption}
\usepackage{subcaption}

\usepackage{algorithm}

\begin{document}
\title{A Universal Non-Parametric Approach For Improved Molecular Sequence Analysis}

\author{Sarwan Ali\inst{1}$^{+,*}$, Tamkanat E Ali\inst{2}$^{+}$, Prakash Chourasia\inst{1}, Murray Patterson\inst{1}}%
% \institute{}
\institute{
$^{1}$Georgia State University, Atlanta, GA, USA \\
$^{2}$Lahore University of Management Science, Lahore, Pakistan
\\
\email{\{sali85, pchourasia1\}@student.gsu.edu, 20100159@lums.edu.pk, mpatterson30@gsu.edu} 
\\
$^{+}$Joint First Authors, * Corresponding author}

\maketitle

\begin{abstract}
In the field of biological research, it is essential to comprehend the characteristics and functions of molecular sequences. The classification of molecular sequences has seen widespread use of neural network-based techniques. Despite their astounding accuracy, these models often require a substantial number of parameters and more data collection. 
In this work, we present a novel approach based on the compression-based Model, motivated from~\cite{jiang2023low}, which combines the simplicity of basic compression algorithms like Gzip and Bz2, with Normalized Compression Distance (NCD) algorithm to achieve better performance on classification tasks without relying on handcrafted features or pre-trained models. Firstly, we compress the molecular sequence using well-known compression algorithms, such as Gzip and Bz2. By leveraging the latent structure encoded in compressed files, we compute the Normalized Compression Distance between each pair of molecular sequences, which is derived from the Kolmogorov complexity. This gives us a distance matrix, which is the input for generating a kernel matrix using a Gaussian kernel. Next, we employ kernel Principal Component Analysis (PCA) to get the vector representations for the corresponding molecular sequence, capturing important structural and functional information. The resulting vector representations provide an efficient yet effective solution for molecular sequence analysis and can be used in ML-based downstream tasks. The proposed approach eliminates the need for computationally intensive Deep Neural Networks (DNNs), with their large parameter counts and data requirements. Instead, it leverages a lightweight and universally accessible compression-based model. Also, it performs exceptionally well in low-resource scenarios, where limited labeled data hinder the effectiveness of DNNs. Using our method on the benchmark DNA dataset, we demonstrate superior predictive accuracy compared to SOTA methods.
\end{abstract}


\section{Introduction}
Molecular sequence analysis stands as a pivotal pursuit in contemporary research, holding the key to unraveling the intricate language encoded in molecular sequences, such as DNA and proteins. The accurate comprehension and classification of these molecular sequences offer profound insights into their structural, functional, and evolutionary characteristics. As the foundation of numerous biological studies, including functional gene annotation, drug discovery, and evolutionary biology, molecular sequence analysis plays an indispensable role in advancing our understanding of the fundamental processes governing life. The pursuit of innovative methodologies in this realm is driven by the quest for more accurate, efficient, and resource-effective approaches to decipher the rich information concealed within the sequences, ultimately contributing to transformative breakthroughs in the broader landscape of molecular biology.

Several methods have been used for the classification of molecular sequences involving Neural Networks, language models, Feature Embedding, and Kernel functions. All these methods face certain challenges when it comes to achieving good accuracy in cases when the available data is less. Neural Network (NN) based methods are one of the most widely employed in molecular sequence classification and have demonstrated impressive accuracy in many cases~\cite{10.1093/bioinformatics/bty22}. However, these methods come with significant limitations including the requirement of a substantial number of parameters and long training times, making them computationally expensive and resource-demanding. Additionally, neural networks and NN-based language models heavily rely on large-scale training data, which may not be readily available for certain biological datasets, particularly in low-resource or rare species scenarios. 

Designing low-dimensional embedding for molecular sequences is a challenging task. One of the most feasible solutions to this challenge is sequence data compression.
% \textcolor{blue}{Tamkanat: this sentence looks a bit weird because the second half of the sentence should come first, shouldn't it be something like: To overcome these limitations we design low-dimensional embedding for molecular sequences based on sequence data compression.}
Some of the well-known compression methods include Gzip, zlib, Bz2~\cite{burrows1994block} etc.
Gzip is used on a large scale for lossless data compression~\cite{10.4108/eai.1-10-2019.160599}, it became popular because of certain characteristics which include being free, open source, robust, compact, portable, has low memory overhead, and has reasonable speed~\cite{KRYUKOV2022100562}. Due to its inertia and its integration with so many sequence analysis tools, even today most of the sequence databases rely on Gzip~\cite{KRYUKOV2022100562}. The zlib and Bz2 compression algorithms efficiently detect non-randomness and low information content~\cite{Vedak_2023}. Their performance gets better as the string length increases. Bz2 compression is used to compress the strings and is not affected by the mass ratios, it does not include character order information due to the process of permuting characters during compression, this negatively affects the accuracy~\cite{jiang2023low}. 
% The LZMA (Lempel–Ziv–Markov chain) algorithm is alignment-free and is used to compress the processed data and capture informative features from the protein sequences~\cite{kalemati2023bicomp}. 
Our compression-based model offers several notable advantages over traditional neural network-based approaches. Firstly, it eliminates the computationally intensive nature of deep neural networks, reducing the parameter requirements and making them more lightweight and accessible. Secondly, by leveraging Gzip compression, our approach can efficiently handle low-resource biological datasets where labeled data is scarce or limited. This enables us to analyze molecular sequences even in resource-constrained scenarios.
Our contributions can be summarized as follows:
\begin{itemize}
    \item We propose a novel approach for analyzing and classifying molecular sequences, using compression-based models including Gzip and Bz2.
    \item We develop an algorithm for Distance Matrix computation, in which we take a set of sequences as input and output a non-symmetric Distance matrix using Normalized Compression Distance (NCD) and different compressors.
    \item We convert the distance matrix into a kernel matrix and extract the low dimensional numerical representation in the end, which can be used as input to any linear and nonlinear machine learning model for supervised and unsupervised analysis. In this way, we also addressed the limitation in~\cite{jiang2023low} where they were only able to apply the k-nearest neighbor classifier for the classification purpose. Hence we show that our proposed method can generalize better for sequence classification.
    \item We also discuss the theoretical justifications for the proposed pipeline including the symmetry of the distance matrix and reproducing Kernel Hilbert Space for the kernel matrix along positive semi-definite, smoothness, continuity, and sensitivity.
    % \item We perform extensive experiments on $3$ real-world datasets and show that in terms of predictive performance, the proposed method outperforms state-of-the-art (SOTA) methods. 
    % \item We also inspect the generated low-dimensional embeddings visually to understand the patterns in the data and show that the proposed embeddings can efficiently group similar classes.
    
\end{itemize}

The rest of the paper is organized as follows: In Section~\ref{sec_rw} we will give details of the literature review followed by the proposed method along with the experimental setup in Section~\ref{sec_PA}.
Our results and experimental evaluation are reported in Section~\ref{sec_EE}. Finally, Section~\ref{sec_conclusion} concludes the paper, emphasizing the effectiveness and potential of our compression-based model in advancing molecular sequence analysis and classification.

\section{Related Work}\label{sec_rw}
Molecular sequence analysis is based on two types of methods, alignment-based and alignment-free. Alignment-based is suitable for small sequences (due to higher dimensionality). The alignment-free method works well for both short and long sequences~\cite{ferragina2007compression}.
Several methods have been used for the analysis of molecular sequences, including neural networks (NN)~\cite{9311619}, language models~\cite{devlin2019bert}, feature embedding~\cite{ali2022pwm2vec}, and kernel functions~\cite{ali2022efficient}. 
% Language modeling (LM) is used to build unsupervised context-based representations of text resulting in improved performance of supervised NLP tasks~\cite{peters1802deep}. 
In recent work, it was seen that the performance of protein prediction tasks can be improved by training language models on protein sequences~\cite{heinzinger2019modeling}. Pretrained language models and word embedding methods have proved to be successful in embedding molecular sequences forming easy-to-process representations that are context-sensitive~\cite{yang2022convolutions}.
In the analysis of big data in proteomics, SeqVec provides a scalable approach for the analysis of the protein data~\cite{heinzinger2019modeling}, which has improved the study of the structure and composition of proteins. 
% SeqVec successfully describes and encodes biochemical properties, but it lacks the context needed to infer certain important aspects which include protein functions~\cite{kane2019augmenting}.
ProtBert is a transformer-based model, that uses a masked language model~\cite{devlin2019bert} it requires positional encoding and has a high memory requirement. 
In the case of NN, several methods have been proposed for sequence analysis~\cite{LE201953}.
% , image-based methods such as chaos game representation~\cite{10.1145/2983468.2983489} and grayscale images~\cite{somodevilla2019dna}. 
Variational AutoEncoder-based methods are also used in the literature for molecular sequence analyses~\cite{9311619}.
These NN-based methods prove to be computationally expensive, face increased risk of overfitting, and are resource-demanding. 
% Their heavy reliance on large training data leads to weak performance in low-resource conditions. 
% \subsection{Feature Embedding methods}
% \textcolor{red}{
% Representation learning methods form fixed-length embeddings for the molecular sequences~\cite{ali2022pwm2vec}. 
In recent works, deep learning-based feature representation methods have been proposed~\cite{shen2018wasserstein,dong2022antimicrobial}.
Several authors proposed feature engineering-based methods to design embeddings for the molecular sequences~\cite{gemci2023deep}.
% }
% \textcolor{blue}{Tamkanat: I want you to please read these three lines and see if you feel a sense of repetition, if so please alter them the way you think would be appropriate} 
Although such methods are efficient in terms of predictive performance, they usually face the problem of the ``curse of dimensionality" due to the higher dimensions of the generated vectors.
Another method used for the sequence analysis is to project the data into high dimensional feature space using kernel matrices~\cite{ghandi2014enhanced,ali2022efficient}.
% The kernel-based learning algorithms have been used in gene expression, prediction of structure and function of molecular sequences. The kernel values for a pair of molecular sequences can be computed based on the matches, mismatches, and gapped $k$-mers (also called nGrams)~\cite{ghandi2014enhanced}. 
However, these methods could cause an overfitting problem along with scalability issues (memory intensive)~\cite{ali2023biosequence2vec}.
Some prominent sequence comparison methods include Normalized Compression Distance (NCD)~\cite{s23031219}, Normalized Information Distance (NID)~\cite{MANTACI2008109}, Euclidean Distance, and Manhattan Distance. The NCD, derived from the concept of Kolmogorov complexity~\cite{kolmogorov1963tables}, provides a measure of similarity between sequences by considering their compressed file sizes. 
% \textcolor{red}{
However, such methods are not used in the literature for representation learning, specifically for molecular sequence analysis.
% }
% \textcolor{blue}{Tamkanat: please go through this line should we keep it as is or does it need to be changed }
% Its use is trending in Biomedical applications, in a recent work~\cite{s23031219}, NCD was used instead of Euclidean Distance and Manhattan Distance as a measure of similarity in forming clusters as part of a newly proposed patient stratification method. NCD allows us to capture important structural information efficiently. Its computable nature makes it a better choice than the Normalized Information Distance (NID) metric~\cite{MANTACI2008109}. 

% \textcolor{blue}{The above behavior shows that objects in the same category share more regularity than objects from different categories.
% The formulation of this concept can be linked to Kolmogorov complexity $Z_s$ and its derived distance metric~\cite{kolmogorov1963tables}. $Z_s$ is the lower bound for measuring information as it represents the length of the shortest binary program that outputs $s$, but there is a limitation that it cannot be used to measure the information content shared between two objects due to its incomputable nature. To overcome this limitation Normalized Compression Distance (NCD) was proposed~\cite{Li-2004} which is computable, it uses compressed length Ls to approximate Kolmogorov complexity $Z_s$. }

\section{Proposed Approach}\label{sec_PA}
We propose an Embedding generation method based on lossless compressors and Normalized Compression Distance (NCD) metric. We start this section by discussing the lossless compression methods below. 

\subsection{Compression Methods}
% \paragraph{\textbf{LZMA compressor:}} 
% The LZMA compressor is based on the Lempel–Ziv–Markov chain algorithm. It is an improved version of LZ coding and is based on constructing a Markov model of the data that captures repeating patterns within it, and then encoding the whole data in terms of this Markov model~\cite{azat-2018}. LZMA can detect repetitions that are far apart in the sequence. This property makes it capable of capturing both intra-sequence and inter-sequence similarities. 
% This compression removes any correlations between the statistical properties of the input data and the output. LZMA compresses repetitive inputs in a manner that the output is free of patterns. 

\paragraph{\textbf{Bz2 Compressor:}}

Bz2 compressor is a general-purpose lossless compressor, based on the Burrows-Wheeler transform (BWT) and Huffman coding~\cite{a13060151}. The BWT is a permutation of the letters of the text (in our case characters/nucleotides of the sequence). After applying BWT to the input, an easily compressible form is generated as it groups symbols into runs of similar units, more precisely the input is divided into blocks of at most $900$ kB, which is compressed separately, keeping in regard to the local similarities in the data. The compression of the transform includes an initial move-to-front encoding with run length encoding and then Huffman encoding.

\paragraph{\textbf{Gzip Compressor:}} 
Gzip uses very few bits to represent information, which is based on the lossless compression algorithm LZ77 (Lempel-Ziv 77 compression algorithm)~\cite{ziv1977universal} and dynamic Huffman algorithm~\cite{10.4108/eai.1-10-2019.160599}. LZ77-Lempel-Ziv compression algorithm encodes a string based on sequential processing. If the present substring was encountered earlier as well then it is encoded with reference to the previous one. A sliding window is used for every new sequence encountered. 
Huffman coding is statistical-based compression, where the symbols are encoded using statistical information such as frequency distribution. There are two types of Huffman coding, dynamic and static. As the data we use is not real-time we use the Dynamic Huffman algorithm, which is a two-pass algorithm. In the first pass, the frequency distribution of symbols is calculated and in the second pass, symbols are encoded. In this technique, depending on the occurrence of the symbols variable length codes are assigned to symbols such that symbols with less occurrence are encoded with more significant bits and symbols with high frequency are encoded with fewer bits, as a result, a good compression ratio is obtained.

\begin{remark}
    Note that our proposed method uses both the compression methods described above separately. 
\end{remark}

\subsection{Problem Formulation}
Given a pair of sequences $s_1$ and $s_2$, where ${s_1,s_2} \in S$ ($S$ is a set of all sequences), we first encode $s_1$ and $s_2$ using UTF-8 encoding~\cite{yergeau1996utf}, which will give us $E_{s_1}$ and $E_{s_2}$.
After encoding, the $E_{s_1}$ and $E_{s_2}$ are compressed using Gzip or Bz2. We will then get the compressed form, denoted by $C_{s_1}$ and $C_{s_2}$. In the next step, we compute the length $L_{s_1}$ and $L_{s_2}$ of the compressed sequences. In a similar way, we compute $L_{s_1 s_2}$, which denotes the length of compressed encoded form for the concatenated sequence $s_1 s_2$. We then use $L_{s_1}$, $L_{s_2}$, and $L_{s_1 s_2}$ as input to Normalized Compression Distance (NCD) approach to get the final distance value, which is calculated using the following expression:

\begin{equation}\label{eq_ncd_ref}
NCD(s_1,s_2)=\frac{Ls_1s_2 - min\{Ls_1,Ls_2\}}{max\{Ls_1,Ls_2\}} 
\end{equation}

% \begin{align}
% {s_1,s_2} \in S \\
% s_1 \neq s_2
% \label{eqn:Input rule}
% \end{align}
In the condition where $s_1 \neq s_2$, the bytes $B$ needed to encode $s_{2}$ based on $s_{1}$ information, i.e. $B_{12}$ can be computed using the following expression:
% represents the number of bytes needed to encode $s_{2}$ based on $s_{1}$ information. 

% $s_{1}$ and $s_{2}$ are part of the input sequences set S. 
\begin{equation}
B_{12} = Ls_1 s_2 - Ls_1 
\end{equation}
% Here $s_{1}$ and $s_{2}$ represent objects belonging to the same category, but a different category from $s_{3}$.
% The $s_{1}s_{2}$ is the concatenated form of $s_{1}$  and $s_{2}$. The notation $L$ represents compressed length,  and 

Similarly, for $s_1$ and $s_3$ sequences, we have

\begin{equation}
B_{13}=Ls_1s_3 - Ls_1 
\end{equation}
% where L$s_{1}s_{3}$ means the compressed length of concatenated version of $s_{1}$ and $s_{3}$. The 
where $B_{13}$ represents the number of bytes needed to encode $s_{3}$ based on $s_1$ information. Given a scenario where $s_1$ and $s_2$ belong to the same category but $s_3$ belongs to a different category than $s_1$ and $s_2$, we have the following expression:

\begin{equation}\label{eq_b12}
B_{12} < B_{13}
\end{equation}

The formulation of the concept mentioned in Equation~\eqref{eq_b12} can be linked to Kolmogorov complexity $Z_s$ (where $s \in S$) and its derived distance metric~\cite{kolmogorov1963tables}. $Z_s$ is the lower bound for measuring information as it represents the length of the shortest binary program that outputs $s$, but there is a limitation that it cannot be used to measure the information content shared between two objects due to the incomputable nature of $Z_s$~\cite{jiang2023low}. 
% \textcolor{blue}{Tamkanat: instead of this can we say that "due to the incomputable nature of $Z_s$" this will make the statement clearer}
To overcome this limitation, Normalized Compression Distance (NCD) is proposed~\cite{Li-2004}, which is computable and uses compressed length $L_s$ to approximate Kolmogorov complexity $Z_s$.

The underlying concept of using compressed length in Equation~\ref{eq_ncd_ref} is that the compressed length is close to $Z_s$. The general rule says, that the higher the compression ratio, the closer $L_s$ is to $Z_s$. Using the NCD-based distance (from Equation~\ref{eq_ncd_ref}), we compute pairwise distances for a set of sequences to generate the required distance matrix.

% The figure~\ref{fig_gzip_overview} shows the overview of the proposed approach.

% \subsection{Distance Matrix computation with Gzip}
\subsection{Our Algorithm}
\begin{figure}[h!]
  \centering
    \includegraphics[scale=0.096]{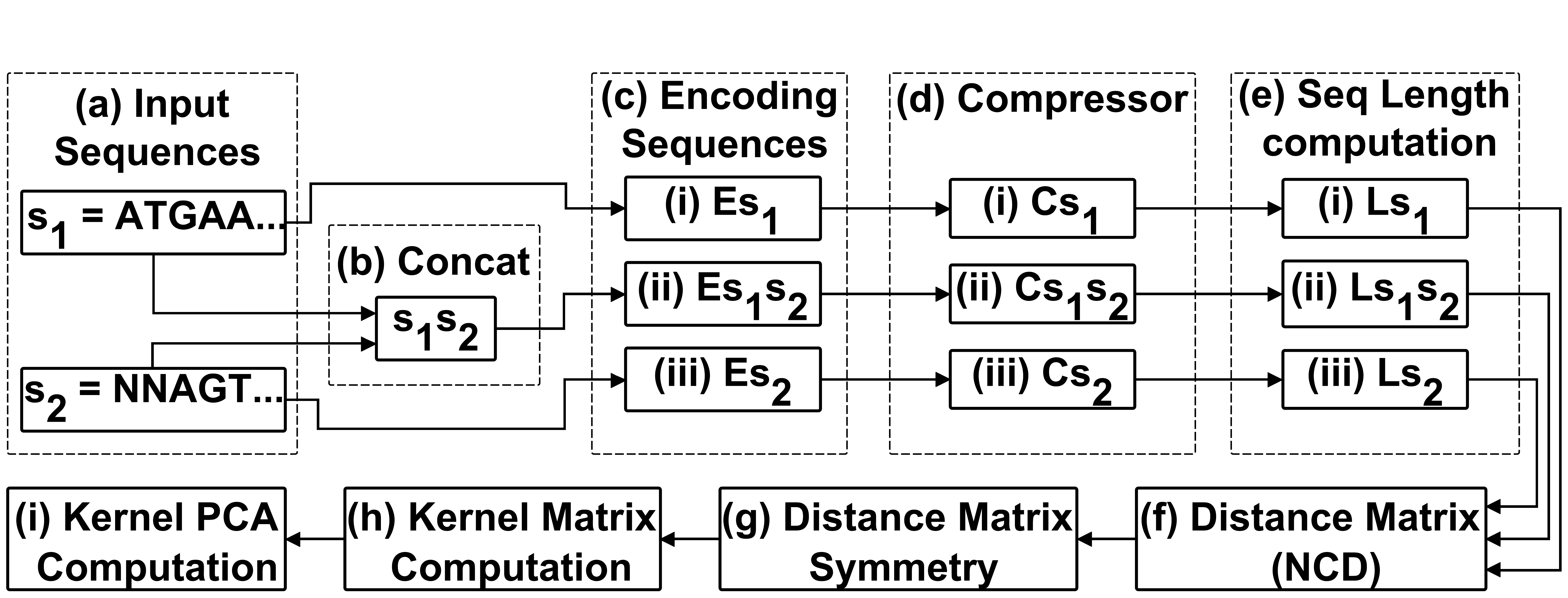}
    \caption{Overview of the proposed approach.}
    \label{fig_gzip_overview}
\end{figure}

In our algorithmic approach (i.e. in Algorithm~\ref{algo_gzip}), we take in a set of sequences (S) as input and output a Distance Matrix (D). We iterate through the Set S, for every sequence referred to as $s$ in our data $S$ and carry out the following steps: 
\begin{enumerate}
  \item Encoded form is generated and stored in a variable E$s_{1}$ (line number 2 of Algorithm~\ref{algo_gzip} and step c(i) of Figure~\ref{fig_gzip_overview})
  \item Encoded E$s_{1}$ is further compressed using Gzip compressor and fed into C$s_{1}$  (line number 3 of Algorithm~\ref{algo_gzip} and step d(i) of Figure~\ref{fig_gzip_overview})
  \item Calculate the length of the compressed C$s_{1}$ and store in a variable referred to as L$s_{1}$.(line number 4 of Algorithm~\ref{algo_gzip} and step e(i) of Figure~\ref{fig_gzip_overview})
\end{enumerate}
To save the Normalized Compression Distance (NCD) between every $s$ and the rest of the sequences in the Set S, we initialize an array termed as D\_ local as shown in the line number 5 of Algorithm~\ref{algo_gzip}.

In another sub-iterative loop, we repeat the steps from 1 to 3 mentioned above for every other sequence in set S (line number 6 of Algorithm~\ref{algo_gzip}). 
% referred to as $s_{2}$ abiding by the condition stated in Equation~\ref{eqn:Input rule}. 
To calculate NCD we first require concatenation of $s_{1}$ and $s_{2}$ (line number 10 of Algorithm~\ref{algo_gzip} and step (b) of Figure~\ref{fig_gzip_overview}), followed by encoding, compression, and calculating its length which is stored in a variable L$s_{1}s_{2}$ (line numbers 11-13 of Algorithm~\ref{algo_gzip} and steps (c)-(e) of Figure~\ref{fig_gzip_overview}). 

Now using the length of the compressed encoded sequences L$s_{1}$, L$s_{2}$ and L$s_{1}s_{2}$, we calculate NCD (line number 14 of Algorithm~\ref{algo_gzip} and step (f) of Figure~\ref{fig_gzip_overview}) and store in the list referred to as D\_ local(line number 15 of Algorithm~\ref{algo_gzip}). At the end of the inner iterative loop, this list is appended in a distance matrix D which would contain the NCD values between every sequence in the set of sequences(S)(line number 17 of Algorithm~\ref{algo_gzip}).

% \textcolor{blue}{SARWAN: Add discussion for Figure~\ref{running_example} here and update its caption. I moved it to supplementary materials.}\textcolor{red}{Tamkanat: Adding it after mentioning figure 1} 
The Figure~\ref{fig_gzip_overview} shows the overview of the proposed approach. 

% A detailed running example of the proposed method is given in Figure~\ref{running_example} in the Appendix.

\begin{algorithm}[h!]
\caption{Distance matrix computation with Gzip}
\label{algo_gzip}
\scriptsize
\begin{algorithmic}[1]

\Statex \textbf{Input:}\texttt{ Set of sequences(S)}
\Statex \textbf{Output:}\texttt{ Distance Matrix(D)}

\For{\texttt{ $s_{1}$ in S\hspace{0.2cm}}}
    \State \texttt{ $Es_{1} \gets encoded \hspace{0.2cm}  s_{1}$}  
    \State \texttt{ $Cs_{1} \gets Gzip\hspace{0.2cm}  compressed \hspace{0.2cm} Es_{1}$} 
    \State \texttt{ $Ls_{1} \gets length \hspace{0.2cm} of \hspace{0.2cm} Cs_{1}$}
    \State \texttt{ $D \_ local \gets [\ ]$}  
    \For{\texttt{ $s_{2}$ in S\hspace{0.2cm}}}
        \State \texttt{ $Es_{2} \gets encoded \hspace{0.2cm}  s_{2}$}  
        \State \texttt{ $Cs_{2} \gets Gzip \hspace{0.2cm} compressed \hspace{0.2cm} Es_{2}$} 
        \State \texttt{ $Ls_{2} \gets length \hspace{0.2cm} of \hspace{0.2cm} Cs_{2}$}  
        \State \texttt{ $s_{1}s_{2} \gets Concatenate(s_{1},s_{2})$}
        \State \texttt{ $Es_{1}s_{2} \gets encoded \hspace{0.2cm}  s_{1}s_{2}$}  
        \State \texttt{ $Cs_{1}s_{2} \gets Gzip \hspace{0.2cm} compressed \hspace{0.2cm} Es_{1}s_{2}$} 
        \State \texttt{ $Ls_{1}s_{2} \gets length \hspace{0.2cm} of \hspace{0.2cm} Cs_{1}s_{2}$}
        \State NCD $\gets$ $\dfrac{L s_1 s_2 - Min (Ls_1, Ls_2)}{Max(Ls_1, Ls_2)}$
        \State \texttt{ $D\_local.append(NCD)$}
        \EndFor
    \State $D.append(D\_local)$
\EndFor
\State return $D$
\end{algorithmic}
\end{algorithm}

\begin{remark}
Our method is better than Deep Neural Networks as there is no need for preprocessing or training, making it simpler. Secondly, fewer parameters with no GPU resources are needed for distance matrix computation, making it lighter, and the absence of underlying assumptions (e.g., assumptions about the data) makes it universal.
\end{remark}

\begin{remark}
    To further understand the idea of NCD-based pairwise distance computation between text/molecular sequences, readers are referred to~\cite{jiang2023low},
\end{remark}

\subsection{Distance Matrix Symmetry}
The Distance matrix (D) obtained using Normalized Compression Distance (NCD) is of size $n \times n$ where $n$ represents the cardinality of the input set $S$. This Distance Matrix D is non-symmetric, so to convert it to symmetric matrix D' we take the average of upper and lower triangle values and replace the original values of the matrix with the average values. 
% as shown in line number of 4 of Algorithm~\ref{algo_symmetric} (in the supplementary materials). 

\subsection{Kernel Matrix Computation}
We generate a Kernel matrix from the symmetric distance matrix (D') of size $n \times n$ using a Gaussian Kernel, where.
\begin{align}
{d'_{ij}}, {d'_{ik}} \in D'
\end{align}

Euclidean Distance ($E$) between two pairs of distances $d'_{ij}$ and $d'_{ik}$ (i.e. distance values computed from pairs of sequences) is calculated using the following equation: 
% \murray{why is it multiplied by 5?  maybe $n$ instead?}
\begin{align}
E_{d'_{ij}, d'_{ik}} = \vert \vert d'_{ij} - d'_{ik} \vert \vert
% \\
% &= ((d'_i1 - d'_j1)^2 + ... + (d'_in - d'_jn)^2) \cdot 5
\end{align} 

The Gaussian Kernel (K) is defined as a measure of similarity between $d'_{ij}$ and $d'_{ik}$. It is represented by the equation below:
\begin{equation} 
K(d'_{ij},d'_{ik})=exp(\frac{-||d'_{ij}-d'_{ik}||^2}{\sigma^2}) 
\end{equation}
where $\sigma^2$ represents the bandwidth of the kernel.  The kernel value is computed as follows:
\begin{numcases}{}
K=1   & if $d'_{ij}$ and $d'_{ik}$ are identical  \\
K\xrightarrow{}0    & if $d'_{ij}$ and $d'_{ik}$ move further apart
\end{numcases}

% K gives us $n$ by $n$ numeric distance matrix that contains the pairwise distances between the rows in D'.
The kernel value is computed for each pair of distances in $D'$ to get the $n \times n$ dimensional kernel matrix.
% \subsection{Kernel PCA-Based Embedding}
Once the kernel matrix is computed, we can leverage kernel Principal Component Analysis (PCA) to derive a lower-dimensional representation of the data. 
% In the context of kernel PCA, this method is extended to handle nonlinear data by employing the kernel matrix to implicitly map the data into a higher-dimensional feature space. In kernel PCA, the principal components are determined based on the eigenvectors of the kernel matrix rather than the original data. This unique approach allows for the discovery of intricate patterns and structures within the data. 
The resulting embeddings, known as kernel principal components, effectively preserve the essential information while retaining the relationships among the molecular sequences including non-linear relations. This representation proves valuable for various downstream tasks, including classification. 
% \textcolor{red}{and visualization.} \textcolor{blue}{Tamkanat: we are not providing visualization should we mention it here} 
% The incorporation of NCD-based distance calculations further enhances the accuracy of the representation and the utilization of the Gaussian kernel facilitates the capture of complex nonlinear patterns, resulting in a more expressive and informative embedding of the molecular sequences. This comprehensive and effective methodology opens new possibilities for analyzing and utilizing molecular sequence data in various bioinformatics applications. A strong theoretical foundation and useful benefits are provided by using the kernel matrix produced from the NCD-based distance matrix to describe molecular sequences. Our compression-based Model performs well in a variety of bioinformatics tasks because it can effectively capture nonlinear relationships and complex patterns.

\subsection{Experimental Setup}\label{sec_ES}
Here we describe dataset statistics and evaluation metrics in detail. 
% Finally, we go over the evaluation metrics that were employed in evaluating the model efficacy.
The experiments are performed on a computer running 64-bit Windows 10 with an Intel(R) Core i5 processor running at 2.10 GHz and 32 GB of RAM. 
% Python is used to implement the code, and for replication, the source code is provided with the published version and is available for online perusal. 
For experiments, we randomly split data into 60-10-30\% for training-validation-testing purposes. The experiments are repeated $5$ times, and we report average results.
Our code is available online for reproducibility~\footnote{\url{https://github.com/sarwanpasha/Non-Parametric-Approach}}.
% \subsection{Dataset Statistics}
% \label{sec_datastats}
We use real-world molecular sequence data comprised of nucleotide sequences. The summary of the dataset used for experimentation is given in Table~\ref{tbl_data_statistics}. 

% A detailed description of each dataset is given in Section~\ref{dataset_description} of the supplementary material.

\begin{table}[h!]
    \centering
    \resizebox{0.8\textwidth}{!}{
    \begin{tabular}{p{1.4cm}ccccclp{6cm}}
    \toprule
    \multirow{2}{1.2cm}{Name} & \multirow{2}{*}{$\vert$ Seq. $\vert$} & \multirow{2}{*}{$\vert$ Classes $\vert$} & \multicolumn{3}{c}{Sequence Statistics} & \multirow{2}{*}{Reference} & \multirow{2}{*}{Description} \\
    \cmidrule{4-6}
        & & & Max & Min & Mean &  &  \\
    \midrule
    \multirow{2}{1.2cm}{Human DNA}  & \multirow{2}{*}{4380} & \multirow{2}{*}{7} & \multirow{2}{*}{18921} & \multirow{2}{*}{5} & \multirow{2}{*}{1263.59} & \multirow{2}{*}{~\cite{human_dna_website_url}}  & Unaligned nucleotide sequences to classify gene family to which humans belong \\
    % \midrule
    %    \multirow{4}{1.2cm}{Coronavirus Host}  & \multirow{4}{*}{5558} & \multirow{4}{*}{21} & \multirow{4}{*}{1584} & \multirow{4}{*}{9} & \multirow{4}{*}{1272.36} & \multirow{4}{1.7cm}{ViPR~\cite{pickett2012vipr}, GISAID~\cite{gisaid_website_url}} &   The spike protein sequences belonging to various clades of the Coronaviridae family accompanied by the infected host label e.g. Humans, Bats, Chickens, etc. \\
    %    \midrule
    %    \multirow{3}{1.2cm}{Spike7k}  & \multirow{3}{*}{7000} & \multirow{3}{*}{22} & \multirow{3}{*}{1274} & \multirow{3}{*}{1274} & \multirow{3}{*}{1274.00} & \multirow{3}{1.5cm}{~\cite{gisaid_website_url}} & The spike protein sequences of the SARS-CoV-2 virus having the information about the coronavirus Lineages of each sequence. \\
     \bottomrule
    \end{tabular}
    }
    \caption{Dataset Statistics. }
    \label{tbl_data_statistics}
\end{table}

% \subsection{Classifiers and Evaluation Metrics}
We used a variety of ML models for classification, including Support Vector Machine (SVM), Naive Bayes (NB), Multi-Layer Perceptron (MLP), K-Nearest Neighbours (KNN), Random Forest (RF), Logistic Regression (LR), and Decision Tree (DT). For performance evaluation, we used average accuracy, precision, recall, F1 (weighted), F1 (macro), Receiver Operator Characteristic Curve (ROC), Area Under the Curve (AUC), and training runtime. 
% We also run our tests by averaging the performance results of $5$ runs for each combination of embedding and classifier to get more consistent results.
% \textcolor{red}{
The baseline models we used for results comparisons include PWM2Vec~\cite{ali2022pwm2vec} which gives each amino acid in the $k$-mers a weight based on where it is located in a $k$-mer position weight matrix (PWM), String Kernel~\cite{ali2022efficient} determines the similarity between the two sequences based on the total number of $k$-mers that are correctly and incorrectly aligned between two sequences, WDGRL~\cite{shen2018wasserstein} a neural network based unsupervised domain adoption technique that uses Wasserstein distance (WD) for feature extraction from input data, Autoencoder~\cite{xie2016unsupervised} that uses a deep neural network to encode data as features which involves iterative optimization of the objective through non-linear mapping from data space X to a smaller-dimensional feature space Z, SeqVec~\cite{heinzinger2019modeling} an ELMO (Embeddings from Language Models) based method for representing biological sequences as continuous vectors, and Protein Bert~\cite{BrandesProteinBERT2022} which is an end-to-end model that does not design explicit embeddings but directly performs classification using a previously learned language model.

\subsection{Justification of Employing the Kernel Matrix}\label{app_justification}

The generation of a kernel matrix from the NCD-based distance matrix offers several technical justifications and significant benefits:

\begin{enumerate}
    \item \textbf{Nonlinearity:} Constructing a kernel matrix based on NCD distances implicitly maps the data into a higher-dimensional feature space, enabling the capture of intricate nonlinear relationships. This is particularly advantageous when dealing with the complex non-linear interactions often present in molecular sequences.
    \item \textbf{Capturing Complex Relationships:} Utilizing the Gaussian kernel in generating the kernel matrix allows for the capture of intricate relationships between sequences. It assigns higher similarity values to similar sequences and lower values to dissimilar ones. This capability enables the representation of complex patterns and structures in the data, surpassing the limitations of linear methods.

\item \textbf{Theoretical Properties of Gaussian Kernel:} The use of the widely adopted Gaussian kernel leverages several underlying theoretical properties. These include the Reproducing Kernel Hilbert Space (RKHS)~\cite{xu2006explicit}, enabling the application of efficient kernel methods like support vector machines. The Universal Approximation property~\cite{hammer2003note} allows the kernel to approximate any continuous function arbitrarily well, making it a powerful tool for modeling complex relationships and capturing nonlinear patterns. Mercer's Theorem~\cite{minh2006mercer} guarantees that the kernel matrix is positive semi-definite, while the smoothness, continuity, and sensitivity to variations further enhance its ability to capture local relationships and adapt to variations in the data distribution.

\item \textbf{Flexibility and Generalization:} The kernel matrix derived from the NCD-based distance matrix can be effectively employed with various machine learning algorithms that operate on kernel matrices. This flexibility allows for the application of a wide range of techniques, including kernel PCA and kernel SVM. Leveraging these algorithms enables the exploitation of the expressive power of the kernel matrix to address diverse tasks, such as dimensionality reduction and classification.

\item \textbf{Preserving Nonlinear Information:} Applying kernel PCA to the kernel matrix captures crucial nonlinear information embedded in the data. This process facilitates the extraction of low-dimensional embeddings that preserve the underlying structure and patterns. Projecting the data onto the principal components retains discriminative information while reducing dimensionality. This proves particularly valuable for handling high-dimensional datasets.
% \textcolor{red}{This proves particularly valuable for visualizing the data} \textcolor{blue}{Tamkanat: we are not visualizing so should we mention it here} and handling high-dimensional datasets.

\item \textbf{Enhanced Performance:} The utilization of the NCD-based distance matrix and kernel matrix can lead to improved performance in various tasks. The incorporation of NCD distance and the capturing of complex relationships in the data enhance the discriminative power of the embeddings. This results in a more accurate classification of the sequences.
% and \textcolor{red}{enhanced visualization of the data} \textcolor{blue}{Tamkanat: Same we are not visualizing so should we mention it here}.
\end{enumerate}

\section{Results And Discussion}\label{sec_EE}
% In this section, we report the classification results for different datasets using average values over $5$ experimental runs. 

% \textcolor{blue}{SARWAN: need to mention why we don't use simple classifiers for protein bert. This is because it is an end-to-end model for classification and does not generate embeddings, unlike other baselines and proposed methods.}

The classification results that are averaged over $5$ runs are reported in Table~\ref{tbl_results_classification_human_dna} for the Human DNA dataset. For the evaluation metrics including average accuracy, precision, recall, weighted and macro F1, and ROC-AUC, our proposed Gzip-based representation outperformed all baselines. For classification training runtime, WDGRL with Naive Bayes performs the best due to the minimum size of embedding compared to other embedding methods. 
% The standard deviation (SD) results for the same dataset and experimental setting (i.e. for $5$ experimental runs) are reported in Table~\ref{tbl_std_org_human_dna} in Section~\ref{SD_results} of the supplementary material. We can observe a pattern that overall the SD values are very small, which essentially means that the results are stable with very small variability.

\begin{table}[h!]
\centering
\resizebox{0.8\textwidth}{!}{
 \begin{tabular}{p{1.9cm}lp{1.1cm}p{1.1cm}p{1.1cm}p{1.7cm}p{1.7cm}p{1.6cm}p{2.5cm}}
    \toprule
        \multirow{1}{1.2cm}{Embeddings} & \multirow{1}{*}{Algo.} & \multirow{1}{*}{Acc. $\uparrow$} & \multirow{1}{*}{Prec. $\uparrow$} & \multirow{1}{*}{Recall $\uparrow$} & \multirow{1}{1.7cm}{F1 (Weig.) $\uparrow$} & \multirow{1}{1.8cm}{F1 (Macro) $\uparrow$} & \multirow{1}{1.6cm}{ROC AUC $\uparrow$} & Train Time (sec.) $\downarrow$\\
        \midrule \midrule	
        \multirow{7}{1.2cm}{PWM2Vec}
        & SVM &  0.302 & 0.241 & 0.302 & 0.165 & 0.091 & 0.505 & 10011.3 \\
         & NB &  0.084 & \underline{0.442} & 0.084 & 0.063 & 0.066 & \underline{0.511} & 4.565 \\
         & MLP &  \underline{0.310} & 0.350 & \underline{0.310} & 0.175 & 0.107 & 0.510 & 320.555 \\
         & KNN &  0.121 & 0.337 & 0.121 & 0.093 & 0.077 & 0.509 & 2.193 \\
         & RF &  0.309 & 0.332 & 0.309 & \underline{0.181} & 0.110 & 0.510 & 65.250 \\
         & LR & 0.304 & 0.257 & 0.304 & 0.167 & 0.094 & 0.506 & 23.651 \\
         & DT &  0.306 & 0.284 & 0.306 & \underline{0.181} & \underline{0.111} & 0.509 & \underline{1.861} \\
        %  \hline
         \cmidrule{2-9} 
        \multirow{7}{1.5cm}{String Kernel}
        & SVM  &  0.618 & 0.617 & 0.618 & 0.613 & 0.588 & 0.753 & 39.791 \\
         & NB   &  0.338 & 0.452 & 0.338 & 0.347 & 0.333 & 0.617 & \underline{0.276} \\
         & MLP  & 0.597 & 0.595 & 0.597 & 0.593 & 0.549 & 0.737 & 331.068 \\
         & KNN  &  0.645 & 0.657 & 0.645 & 0.646 & 0.612 & 0.774 & 1.274 \\
         & RF   &  \underline{0.731} & \underline{0.776} & \underline{0.731} & \underline{0.729} & \underline{0.723} & \underline{0.808} & 12.673 \\
         & LR   &  0.571 & 0.570 & 0.571 & 0.558 & 0.532 & 0.716 & 2.995 \\
         & DT   & 0.630 & 0.631 & 0.630 & 0.630 & 0.598 & 0.767 & 2.682 \\
          \cmidrule{2-9} 
           \multirow{7}{1.2cm}{WDGRL}  
             & SVM & 0.318 & 0.101 & 0.318 & 0.154 & 0.069 & 0.500 & 0.751 \\
             & NB & 0.232 & 0.214 & 0.232 & 0.196 & 0.138 & 0.517 & \textbf{\underline{0.004}} \\
             & MLP &  0.326 & 0.286 & 0.326 & 0.263 & 0.186 & 0.535 & 8.613 \\
             & KNN & 0.317 & 0.317 & 0.317 & 0.315 & 0.266 & 0.574 & 0.092 \\
             & RF & \underline{0.453} & \underline{0.501} & \underline{0.453} & \underline{0.430} & \underline{0.389} & \underline{0.625} & 1.124 \\
             & LR & 0.323 & 0.279 & 0.323 & 0.177 & 0.095 & 0.507 & 0.041 \\
             & DT & 0.368 & 0.372 & 0.368 & 0.369 & 0.328 & 0.610 & 0.047 \\
         \cmidrule{2-9} 
            \multirow{7}{1.5cm}{Autoencoder}
             & SVM & 0.621 & 0.638 & 0.621 & 0.624 & 0.593 & 0.769 & 22.230 \\
             & NB &  0.260 & 0.426 & 0.260 & 0.247 & 0.268 & 0.583 & 0.287 \\
             & MLP & 0.621 & 0.624 & 0.621 & 0.620 & 0.578 & 0.756 & 111.809 \\
             & KNN & 0.565 & 0.577 & 0.565 & 0.568 & 0.547 & 0.732 & \underline{1.208} \\
             & RF & 0.689 & 0.738 & 0.689 & 0.683 & 0.668 & 0.774 & 20.131 \\
             & LR & \underline{0.692} & \underline{0.700} & \underline{0.692} & \underline{0.693} & \underline{0.672} & \underline{0.799} & 58.369 \\
             & DT & 0.543 & 0.546 & 0.543 & 0.543 & 0.515 & 0.718 & 10.616 \\
              \cmidrule{2-9} 
            \multirow{7}{1.5cm}{SeqVec}
             & SVM & 0.656 & 0.661 & 0.656 & 0.652 & 0.611 & \underline{0.791} & 0.891 \\
             & NB & 0.324 & 0.445 & 0.312 & 0.295 & 0.282 & 0.624 & \underline{0.036} \\
             & MLP & 0.657 & 0.633 & 0.653 & 0.646 & 0.616 & 0.783 & 12.432 \\
             & KNN & 0.592 & 0.606 & 0.592 & 0.591 & 0.552 & 0.717 & 0.571 \\
             & RF & 0.713 & \underline{0.724} & 0.701 & 0.702 & \underline{0.693} & 0.752 & 2.164 \\
             & LR & \underline{0.725} & 0.715 & \underline{0.726} & \underline{0.725} & 0.685 & 0.784 & 1.209 \\
             & DT & 0.586 & 0.553 & 0.585 & 0.577 & 0.557 & 0.736 & 0.24 \\
             \cmidrule{2-9} 
            \multirow{1}{1.9cm}{Protein Bert}
             & \_ & 0.542 & 0.580 & 0.542 & 0.514 & 0.447 & 0.675 & 58681.57 \\
              \cmidrule{2-9} 
            \multirow{7}{1.5cm}{Gzip (ours)} 
             & SVM & 0.692 & 0.844 & 0.692 & 0.699 & 0.692 & 0.771 & 2.492 \\
             & NB & 0.464 & 0.582 & 0.464 & 0.478 & 0.472 & 0.704 & \underline{0.038} \\
             & MLP & \underline{\textbf{0.831}} & 0.833 & \underline{\textbf{0.831}} & \underline{\textbf{0.830}} & \underline{\textbf{0.813}} & \underline{\textbf{0.890}} & 7.546 \\
             & KNN & 0.773 & 0.792 & 0.773 & 0.776 & 0.768 & 0.856 & 0.193 \\
             & RF & 0.810 & \underline{\textbf{0.858}} & 0.810 & 0.812 & 0.811 & 0.858 & 6.539 \\
             & LR & 0.621 & 0.822 & 0.621 & 0.616 & 0.581 & 0.712 & 0.912 \\
             & DT & 0.648 & 0.651 & 0.648 & 0.648 & 0.624 & 0.780 & 2.590 \\
             \cmidrule{2-9} 
            \multirow{7}{1.5cm}{Bz2 (ours)} 
             & SVM & 0.545 & 0.769 & 0.545 & 0.524 & 0.501 & 0.669 & 2.856 \\
             & NB & 0.403 & 0.577 & 0.403 & 0.411 & 0.410 & 0.653 & \underline{0.034} \\
             & MLP & 0.696 & 0.702 & 0.696 & 0.698 & 0.670 & 0.809 & 7.601 \\
             & KNN & 0.697 & 0.715 & 0.697 & 0.699 & 0.677 & \underline{0.813} & 0.215 \\
             & RF & \underline{0.720} & \underline{0.804} & \underline{0.720} & \underline{0.722} & \underline{0.721} & 0.798 & 6.000 \\
             & LR & 0.488 & 0.721 & 0.488 & 0.449 & 0.401 & 0.626 & 0.899 \\
             & DT & 0.574 & 0.577 & 0.574 & 0.574 & 0.547 & 0.735 & 2.290 \\
            %  \cmidrule{2-9} 
            % \multirow{7}{1.5cm}{LZMA (ours)} 
            %  & SVM & 0.660 & 0.833 & 0.660 & 0.662 & 0.653 & 0.747 & 2.388 \\
            %  & NB & 0.488 & 0.604 & 0.488 & 0.498 & 0.506 & 0.718 & \underline{0.036} \\
            %  & MLP & \underline{\textbf{0.859}} & 0.862& \underline{\textbf{0.859}} & \underline{\textbf{0.860}} & \underline{\textbf{0.838}} & \underline{\textbf{0.910}} & 4.711 \\
            %  & KNN & 0.750 & 0.769& 0.750 & 0.753 & 0.743 & 0.845 & 0.165 \\
            %  & RF & 0.819 & \underline{\textbf{0.864}} & 0.819 & 0.821 & 0.821 & 0.864 & 6.204 \\
            %  & LR & 0.591 & 0.809 & 0.591 & 0.582 & 0.552 & 0.694 & 0.996 \\
            %  & DT & 0.644 & 0.646 & 0.644 & 0.644 & 0.617 & 0.778 & 2.434 \\

         \bottomrule
         \end{tabular}
}
 \caption{Classification results (averaged over $5$ runs) on \textbf{Human DNA} dataset for different evaluation metrics. The best classifier performance for every embedding is shown with the underline. Overall best values are shown in bold.}
    \label{tbl_results_classification_human_dna}
\end{table}

To test the statistical significance of results, we used the student t-test and observed the $p$-values using averages and SD results of $5$ runs. We observed that the SD values for all datasets and metrics are very small i.e. mostly $<0.002$, we also noted that $p$-values were $< 0.05$ in the majority of the cases (because SD values are very low). This confirmed the statistical significance of the results. 
% The proof of statistical significance further made our claim stronger, where we say that the parameter-free representation using the NCD-based compression using methods like Gzip, Bz2, and LZMA outperforms the baselines and state-of-the-art methods (such as pre-trained Protein Bert and Large Language Model-based SeqVec) in terms of supervised analysis.

% \subsection{Discussion}
From the overall average classification results, SD results, and statistical significance results, we can conclude that the proposed NCD compression-based method can outperform the SOTA for predictive performance on real-world molecular sequence dataset.
Moreover, even after fine-tuning the Large language model (LLM) such as SeqVec, the proposed parameter-free method significantly outperforms the LLM for all evaluation metrics. With the theoretical justifications and statistical significance of the results, we can conclude that using the proposed method in a real-world scenario for molecular sequence analysis can help biologists understand different viruses and deal with future pandemics efficiently.

\section{Conclusion}\label{sec_conclusion}
In conclusion, we propose lightweight and efficient compression-based models for classifying molecular sequences. By combining the simplicity of the compression methods (e.g., Gzip and Bz2) with a powerful nearest neighbor algorithm, our method achieves state-of-the-art performance without the need for extensive parameter tuning or pre-trained models. 
The compression-based Model successfully overcomes the limitations of neural network-based methods, offering a more accessible and computationally efficient solution, especially in low-resource scenarios. 
% Through evaluations of benchmark datasets, we have demonstrated the model's superior predictive accuracy compared to existing methods. 
In the future, we will be exploring the applications of our model in other bioinformatics domains and investigating ways to further optimize and tailor the approach for specific biological datasets.

\bibliographystyle{IEEEtran}
\bibliography{references}

\end{document}